\documentclass[letterpaper]{article} 
\usepackage{aaai2026}  
\usepackage{times}  
\usepackage{helvet}  
\usepackage{courier}  
\usepackage[hyphens]{url}  
\usepackage{graphicx} 
\usepackage{natbib}  
\usepackage{caption} 
\usepackage{algorithm}
\usepackage{algorithmic}
\usepackage{multirow}
\usepackage{makecell}

\usepackage[utf8]{inputenc} 
\usepackage{amsmath}        
\usepackage{amssymb}        
\DeclareUnicodeCharacter{2212}{-} 

\usepackage{newfloat}
\usepackage{listings}
\DeclareCaptionStyle{ruled}{labelfont=normalfont,labelsep=colon,strut=off} 
\floatstyle{ruled}
\newfloat{listing}{tb}{lst}{}
\floatname{listing}{Listing}
\title{Multi-granularity Interactive Attention Framework for Residual Hierarchical Pronunciation Assessment}
\author{
    Hong Han, Hao-Chen Pei, Zhao-Zheng Nie, Xin Luo, Xin-Shun Xu\thanks{Corresponding Author}\\
}
\affiliations{
    School of Software, Shandong University, Jinan, China\\

    sparkinhan@163.com, \{202235343, 202435350\}@mail.sdu.edu.cn, 
    \\luoxin.lxin@gmail.com, xuxinshun@sdu.edu.cn
}

\usepackage{bibentry}

\begin{document}

\maketitle

\begin{abstract}
Automatic pronunciation assessment plays a crucial role in computer-assisted pronunciation training systems. Due to the ability to perform multiple pronunciation tasks simultaneously, multi-aspect multi-granularity pronunciation assessment methods are gradually receiving more attention and achieving better performance than single-level modeling tasks. However, existing methods only consider unidirectional dependencies between adjacent granularity levels, lacking bidirectional interaction among phoneme, word, and utterance levels and thus insufficiently capturing the acoustic structural correlations. To address this issue, we propose a novel residual hierarchical interactive method, HIA for short, that enables bidirectional modeling across granularities. As the core of HIA, the Interactive Attention Module leverages an attention mechanism to achieve dynamic bidirectional interaction, effectively capturing linguistic features at each granularity while integrating correlations between different granularity levels. We also propose a residual hierarchical structure to alleviate the feature forgetting problem when modeling acoustic hierarchies. In addition, we use 1-D convolutional layers to enhance the extraction of local contextual cues at each granularity. Extensive experiments on the speechocean762 dataset show that our model is comprehensively ahead of the existing state-of-the-art methods.
\end{abstract}


\section{Introduction}
In the field of language learning, computer-assisted pronunciation training system (CAPT) \cite{eskenazi_Overview-Spoken-Language_2009, tejedor-garcia_Assessing-Pronunciation-Improvement_2020}, utilizing computer technology to assist language learners in improving their pronunciation skills, provides interactive training methods with immediate feedback. As the core component of CAPT, automatic pronunciation assessment (APA) \cite{li_Intonation-Classification-L2_2017, kheir_Automatic-Pronunciation-Assessment_2023} aims to rate the quality of a speaker's pronunciation and provides detailed feedback to better assist foreign language learning. Early researches on APA tend to be centered around signal granularity of speech data, such as assessing pronunciation accuracy at phoneme level \cite{wang2012improved} or detecting various aspect at word or utterance levels \cite{tepperman_Automatic-Syllable-Stress_2005, arias_Automatic-Intonation-Assessment_2010}. These single-granularity assessment methods perform well in some specific tasks they are designed to address, but they have many limitations. In particular, they do not take the natural complexity and multi-granularity nature of speech into account \cite{lin_Automatic-Scoring-MultiGranularity_2020}.

\begin{figure}[t]
\centering
\includegraphics[width=3.2 in]{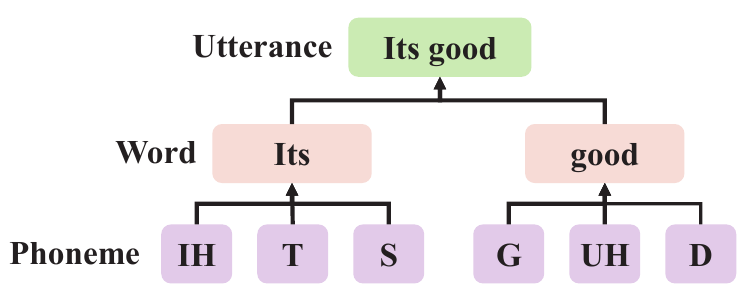}
\caption{\small Schematic diagram of the acoustic hierarchical structure with a sample utterance "Its good".}
\label{fig_1}
\end{figure}

The granularities among the pronunciation assessment tasks are not separated from each other \cite{cincarek_Automatic-Pronunciation-Scoring_2009}, and they have some implicit correlations as shown in Fig. \ref{fig_1}. Acoustic signals are typically characterized by their intricate hierarchical structure, with pronunciation results at lower granularity levels affecting higher granularity levels \cite{al2014pronunciation}. However, modeling a single granularity level cannot fully reveal this implicit relations between different granularity levels.

Recently, to comprehensively study acoustic features at multiple levels of granularity in read-aloud scenario, research endeavors integrate multi-aspect multi-granular pronunciation assessment tasks into a single model to simultaneously evaluate multiple aspects of pronunciation including accuracy, fluency, prosody, and completeness within a unified model across different granularities (i.e., phoneme, word, and utterance).

However, existing methods have some limitations. GOPT \cite{gong_TransformerBased-MultiAspect-MultiGranularity_2022} can effectively handle different granularity scoring tasks when modeling multi-granularity tasks in parallel, but lacks interaction between granularities, which may restrict the modeling of complex correlations between different granularities. HiPAMA \cite{do_Hierarchical-Pronunciation-Assessment_2023} uses a hierarchical structure to capture granularity dependencies, but its information flow is unidirectional, failing to consider bidirectional interaction. Gradformer \cite{pei_Gradformer-Framework-MultiAspect_2024} focuses on utterance modeling and fails to capture the correlations between phoneme and word levels. HierGAT \cite{yan_Effective-Hierarchical-Graph_2024} uses graph neural networks for hierarchical modeling, but its fixed graph structure limits the dynamic interaction between different granularity levels. As mentioned above, these methods only consider unidirectional relations between adjacent granularities, such as how phonemes form word pronunciations, and lack interactive modeling among phoneme, word, and utterance levels, failing to achieve bidirectional interaction. Additionally, for hierarchical modeling methods, as the granularity level increases, the corresponding model depth also increases, which may lead to the forgetting of initial encoded features.

Bidirectional interaction between different granularities is crucial \cite{gao2022paraformer}. For example, the same word may be stressed differently depending on the utterances in English. The lack of modeling for this pronunciation pattern may be the reason why previous methods perform poorly on word stress.

To address the aforementioned issues, we propose a new residual hierarchical interactive multi-aspect multi-granular pronunciation assessment framework, HIA. Specifically, we design an interactive attention module that enables bidirectional interaction at each granularity level. This module processes the features of each granularity in the acoustic embeddings through the attention mechanism and generates interactive attention heads for each granularity to effectively capture the correlations between different granularities, thereby achieving the bidirectional interaction between granularity levels. Additionally, HIA optimizes the hierarchical structure using a residual connection \cite{he_Deep-Residual-Learning_2016}, i.e., introducing acoustic embeddings from the Transformer encoder when modeling the target granularity. By adopting the residual structure, we alleviate the forgetting and processing limitations of the original embedding features caused by the increased depth of the model.

Contributions of this paper are summarized as follows:

\begin{itemize}
\item{We first note that prior methods perform poorly on word stress, as the same word can be stressed differently across utterances in English, and then introduce the HIA framework to address this limitation.}
\item{To address the issue of insufficient inter-granularity interaction, we design an interactive attention module to enable bidirectional interaction across phoneme, word, and utterance levels, thereby capturing their correlations more effectively and overcoming prior interaction limitations.}
\item{To alleviate the feature forgetting in hierarchical modeling, we propose a residual hierarchical structure, which allows HIA to effectively leverage the hierarchical structure characteristics of speech signals while mitigating the forgetting of initial encoding features by the hierarchical structure, thereby improving the overall performance of the model}
\item{We conduct extensive experiments and analyses on the speechocean762 dataset, experimental results show that our model achieves state-of-the-art performance on all metrics.}
\end{itemize}


\section{Related Work}
\label{section2}
As the core technology of CAPT research, pronunciation assessment can be simply divided into two categories according to task scenarios: open-response pronunciation assessment and read-aloud pronunciation assessment.

\subsection{Open-response Pronunciation Assessment}
Open-response pronunciation assessment demands the system to handle learners' spontaneous pronunciation without pre-specified texts, making it particularly critical in open-response scenarios, such as IELTS. In these scenarios, learners must accomplish free or semi-free pronunciation tasks through oral expression, which poses higher demands on speech assessment technology.


In this field, the MultiPA \cite{chen_MultiPA-Multitask-Speech_2024} represents a significant advancement. In concrete terms, the model leverages pre-trained self-supervised learning models and Automatic Speech Recognition (ASR) models to identify potential words. In addition, researchers are also exploring methods for scoring that do not rely on ASR. Cheng et al. (\citeyear{cheng_ASRFree-Pronunciation-Assessment_2020a}) investigated an ASR-free scoring approach that is derived from the marginal distribution of raw speech signals. Cheng et al. (\citeyear{liu_ASRFree-Fluency-Scoring_2023}) proposed a novel ASR-free approach for automatic fluency assessment using self-supervised learning.


\subsection{Read-aloud Pronunciation Assessment}
Unlike open-response pronunciation assessment, in read-aloud pronunciation assessment tasks, learners are required to read pre-specified text in a read-aloud scenario. 

In early researches on APA, Witt et al. (\citeyear{witt_Phonelevel-Pronunciation-Scoring_2000}) proposed a phoneme-level pronunciation scoring and evaluation method to derive the posterior probabilities of phonemes, thus assessing the "Goodness of Pronunciation" (GOP). Hu et al. (\citeyear{hu_Improved-Mispronunciation-Detection_2015}) enhanced mispronunciation detection and diagnosis (MDD) \cite{strik_Comparing-Different-Approaches_2009, li_Mispronunciation-Detection-Diagnosis_2017} by employing an acoustic model trained with deep neural networks and a transfer-learning-based logistic regression classifier. Although such approaches are interpretable, they implicitly assumed that different granularity levels are independent which leads to suboptimal performance.

\begin{figure*}[t]
\centering
\includegraphics[width=7 in]{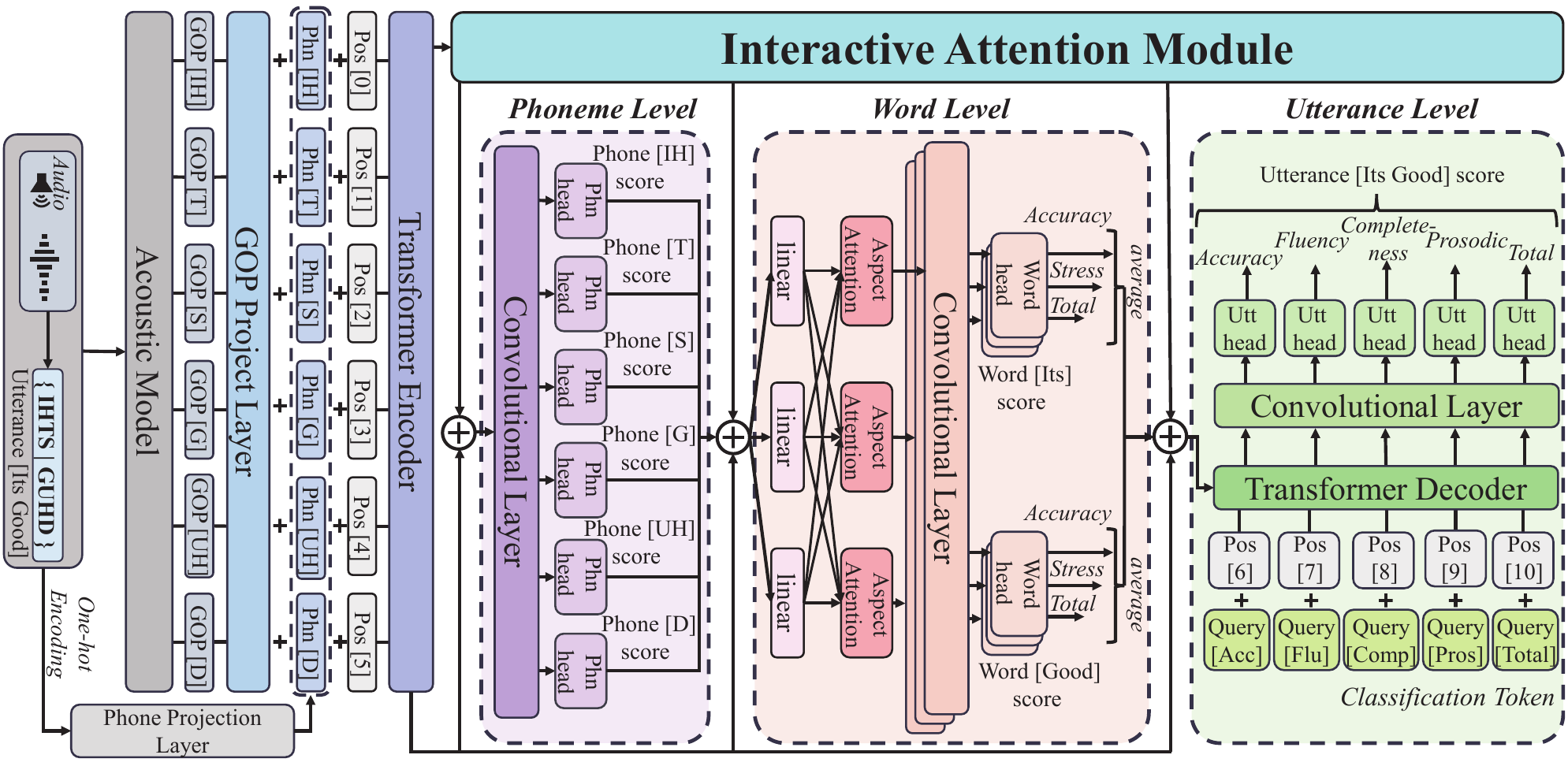}
\caption{Main architecture of HIA. HIA takes the GOP features extracted from the acoustic model and the projected canonical phoneme embeddings as input. Then, Transformer encoder is applied to encode the input to obtain the acoustic embeddings. Finally, the integrated residual hierarchical structure is used to obtain the scores at each granularity in turn.}
\label{fig_2}
\end{figure*}

With breakthroughs in neural network architectures and optimization algorithms \cite{vaswani_Attention-All-You_2017, gao2017video}, the research has shifted towards multi-aspect Multi-granularity pronunciation assessment. A notable research of this transition is the GOPT \cite{gong_TransformerBased-MultiAspect-MultiGranularity_2022}, which introduces an innovative Transformer-based multi-task learning framework, achieving better results than a single-task-specific assessment task. Building on GOPT, Do et al. (\citeyear{do_Hierarchical-Pronunciation-Assessment_2023}) proposed the HiPAMA, which adopts hierarchical structure to sequentially assess pronunciation at various granularity levels. Furthermore, Pei et al. (\citeyear{pei_Gradformer-Framework-MultiAspect_2024}) introduced the Gradformer with granularity-decoupled structure, which incorporates a convolution-enhanced Transformer encoder to encode acoustic features. In addition to GOP based methods, several studies have used non-GOP methods such as transfer learning and self-supervised learning \cite{kim_Automatic-Pronunciation-Assessment_2022, chao_Hierarchical-Contextaware-Modeling_2023, lin_Exploiting-Information-Native_2023} to cope with limited L2 training data.



\section{Methodology}
\label{section3}
\subsection{Overview}
As shown in Fig. \ref{fig_2}, our model adopts residual structure, namely utilizes acoustic feature embeddings initially encoded by the Transformer encoder for each granularity. These embeddings are combined with the output of the interactive attention heads for each granularity. For word- and utterance-level granularity, we also incorporate scoring results from the phoneme- and word-level, respectively, to model the hierarchical structure. Each component is detailed in the following subsections.

\subsection{Acoustic Feature Processing}

For fair comparison, we follow the baseline model \cite{gong_TransformerBased-MultiAspect-MultiGranularity_2022} to use GOP features \cite{tu_Investigating-Role-L1_2018b, shi2020context} as input to the model. In our experiments, ASR acoustic model is used to extract GOP feature which is the log phone posterior (LPP) and log posterior ratio (LPR) defined in \cite{hu_Improved-Mispronunciation-Detection_2015}. Specifically, the LPP of a phone ${p}$ is defined as follows:
\begin{equation}
P(p|o_t) = \sum_{s \in p}P(s|o_t),
\end{equation}
\begin{equation}
LPP(p) \approx \frac{1}{t_e-t_s+1}\sum_{t=t_s}^{t_e} \log P(p|o_t),
\end{equation}
where ${o_t}$ is the input observation of the frame ${t}$, ${s}$ is the state belonging to the phone ${p}$; ${t_s}$ and ${t_e}$ are the start and end frame indexes, respectively. LPR of a phone ${p_j}$ versus ${p_i}$ is defined as:
\begin{equation}
LPR(p_j|p_i) = \log P(p_j|\textbf{o};t_s,t_e)-\log P(p_i|\textbf{o};t_s,t_e).
\end{equation}

The Librispeech \cite{panayotov_Librispeech-ASR-Corpus_2015} acoustic model we use to process audio and generate forced alignment has a total of 42 pure phones, thus the GOP feature of phone p can be defined as an 84-dimensional vector as follows:
\begin{equation}
[LPP(p_1),...,LPP(p_{42}),LPR(p1|p),...,LPR(p_{42}|p)].
\end{equation}

Considering that different phonemes exhibit distinct characteristics, we use the canonical phoneme embedding to provide useful information same as the baseline model \cite{gong_TransformerBased-MultiAspect-MultiGranularity_2022}. Then, we add the projected GOP feature, canonical phoneme embedding, and a trainable positional embedding together and input them to the Transformer encoder.

\subsection{Interactive Attention Module}
\begin{figure}[t]
\centering
\includegraphics[width=3.3 in]{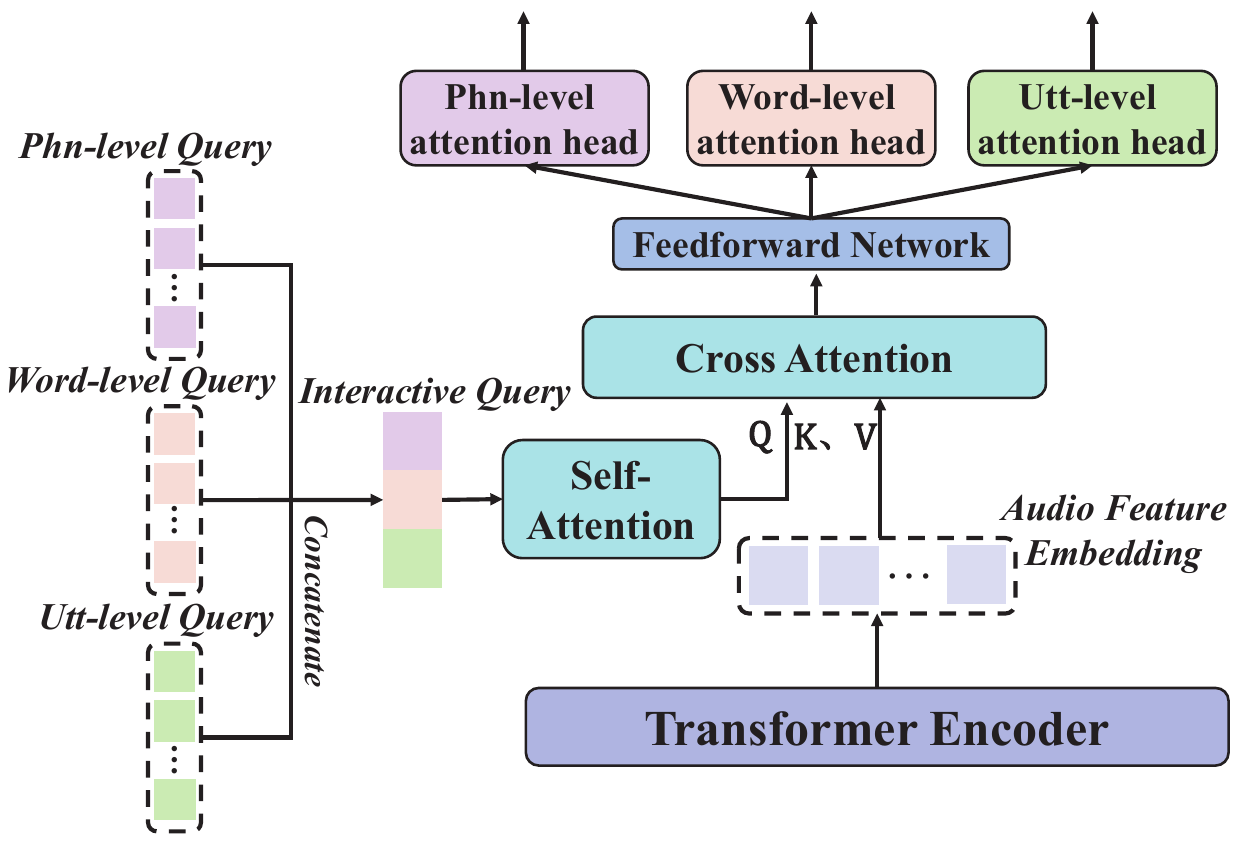}
\caption{Network structure of interactive attention module. For simplicity, the residual connection and norm layers are omitted. Phn is Phoneme, Utt is Utterance.}
\label{fig_3}
\end{figure}

In the field of multi-aspect multi-granularity pronunciation assessment, effectively leveraging correlations among granularities is critical for accurately predicting pronunciation scores. Previous studies have only considered unidirectional relations between adjacent granularities (i.e., phoneme $\rightarrow$ word $\rightarrow$ utterance), thereby neglecting the bidirectional correlations between multiple granularities. For the first time, we introduce an Interactive Attention Module that jointly encodes all pairwise bidirectional interaction within a single self-attention operation, thereby enabling simultaneous bottom-up and top-down information exchange across phoneme, word, and utterance levels as shown in Fig. \ref{fig_3}.

First, we initialize a set of query vectors for each granularity by projecting acoustic feature embeddings, referred to as ${Q^l \in \mathbb{R}^{{B} \times {D}}}$, here $l$ represents different granularity level, $B$ is the batch size, and $D$ is the embedding dimensions for each granularity respectively. These queries represent the initial states of different granularities. Subsequently, we concatenate the multi-granularity queries as ${Q=\{Q^{phn}, Q^{word}, Q^{utt}\}}$, ${Q\in \mathbb{R}^{{B} \times {3} \times {D}}}$. Then, the self-attention mechanism is applied to $Q$ not only enables bidirectional interaction between different granularity levels but also effectively captures the correlations within each granularity level, generating self-attention heads for concatenated multi-granularity query as follows:
\begin{equation}
Q_{self} = SelfAttn(Q).
\end{equation}

By concatenating multi-granularity queries and introducing the self-attention mechanism, the model achieves bidirectional interaction between different granularity levels. The generated self-attention heads not only contain multi-granularity bidirectional interaction features but also preserve each level’s native cues, further improving the model's performance in multi-granularity tasks.

Subsequently, we input self-attention heads together with the pre-processed acoustic feature embeddings, into the cross-attention mechanism. The self-attention heads $Q_{self}$ serve as queries, while the acoustic feature embeddings ${X}$ act as keys and values, thereby mapping multi-granularity interaction features to the acoustic feature space. The operation is formulated as follows:
\begin{equation}
Q_{cross} = CrossAttn(Q_{self}, X).
\end{equation}


Finally, the output of the cross-attention mechanism is fed into a feed forward network and the output $H$ is projected to obtain interactive attention heads for each granularity, denoted as ${H^{phn}}$, ${H^{word}}$ and ${H^{utt}}$. Subsequently, the interactive attention heads for each granularity are used in the corresponding granularity's modeling process, enabling more precise scoring in multi-granularity pronunciation assessment tasks.


\subsection{Residual Hierarchical Multi-granularity Modeling}
\subsubsection{Phoneme-level Modeling}
The output $X$ of the Transformer encoder is added to the phoneme-level attention heads in the interactive attention module. Finally, the fused features are input into the convolutional layer, which further extracts and refines the phoneme-level features by learning the characteristic patterns of local regions \cite{abdel2014convolutional}. 

As shown in Fig. \ref{fig_2}, we add phoneme-level regression heads after the output of each corresponding phoneme in the convolutional layer. Thereinto, each phoneme has a phoneme-level regression head, a ${48\times1}$ linear layer with layer normalization, that outputs phoneme-level accuracy scores. The model outputs phoneme-level scores that reflect the learner's pronunciation quality in terms of phoneme accuracy. The formula modeling process is as follows: 
\begin{equation}
S^{phn} = Conv(X + H^{phn}).
\end{equation}

\begin{table*}[t]
    \setlength{\tabcolsep}{1mm} 
    \centering
    \small
        \begin{tabular}{c c|c c|c c c|c c c c c}
            \hline
            \multicolumn{2}{c|}{\multirow{2}{*}{\textbf{Model}}} &\multicolumn{2}{c|}{\textbf{Phoneme score}} &\multicolumn{3}{c|}{\textbf{Word score (PCC)}} &\multicolumn{5}{c}{\textbf{Utterance score (PCC)}} \\
            \cline{3-12}
            \multicolumn{2}{c|}{} &MSE$\downarrow$ &PCC$\uparrow$ &Acc$\uparrow$ &Stress$\uparrow$ &Total$\uparrow$ &Acc$\uparrow$ &Comp$\uparrow$ &Fluency$\uparrow$ &Prosodic$\uparrow$ &Total$\uparrow$ \\
            \hline
            \multicolumn{2}{l|}{Human} &- &0.555 &0.589 &0.212 &0.602 &0.618 &0.658 &0.665 &0.651 &0.675\\
            \hline
            \multicolumn{2}{l|}{RF \cite{zhang_Speechocean762-OpenSource-NonNative_2021a}} &0.130 &0.440 &- &- &- &- &- &- &- &-\\
            \multicolumn{2}{l|}{SVR \cite{zhang_Speechocean762-OpenSource-NonNative_2021a}} &0.160 &0.450 &- &- &- &- &- &- &- &-\\
            \multicolumn{2}{l|}{UOR \cite{mao_Universal-Ordinal-Regression_2022}} &0.120 &0.520 &- &- &- &- &- &- &- &-\\
            \multicolumn{2}{l|}{Mixup-pretrain \cite{fu_Improving-Nonnative-Wordlevel_2022}} &- &- &- &- &0.610 &- &- &- &- &-\\
            \multicolumn{2}{l|}{Deep feature \cite{lin_Deep-Feature-Transfer_2021}} &- &- &- &- &- &- &- &- &- &0.720\\
            \multicolumn{2}{l|}{Wav2vec2-based \cite{lin_Exploiting-Information-Native_2023}} &- &- &- &- &- &- &- &- &- &0.725\\
            \multicolumn{2}{l|}{LAS \cite{liu_Leveraging-PhoneLevel-LinguisticAcoustic_2023}} &- &- &- &- &- &- &- &- &- &\textbf{0.766}\\
            \hline
            \multicolumn{2}{l|}{\multirow{2}{*}{LSTM \cite{gong_TransformerBased-MultiAspect-MultiGranularity_2022}}} &0.089 &0.591 &0.514 &0.294 &0.531 &0.720 &0.076 &0.045 &0.747 &0.741\\
            \multicolumn{2}{l|}{} &$\pm$0.000 &$\pm$0.003 &$\pm$0.003 &$\pm$0.012 &$\pm$0.004 &$\pm$0.002 &$\pm$0.086 &$\pm$0.002 &$\pm$0.005 &$\pm$0.002\\
            \hline
            \multicolumn{2}{l|}{\multirow{2}{*}{GOPT \cite{gong_TransformerBased-MultiAspect-MultiGranularity_2022}}} &0.085 &0.612 &0.533 &0.291 &0.549 &0.714 &0.155 &0.753 &0.760 &0.742\\
            \multicolumn{2}{l|}{} &$\pm$0.001 &$\pm$0.003 &$\pm$0.004 &$\pm$0.030 &$\pm$0.002 &$\pm$0.004 &$\pm$0.039 &$\pm$0.008 &$\pm$0.006 &$\pm$0.005\\
            \hline
            \multicolumn{2}{l|}{\multirow{2}{*}{HiPAMA \cite{do_Hierarchical-Pronunciation-Assessment_2023}}} &0.084 &0.616 &0.575 &0.320 &0.591 &0.730 &0.276 &0.749 &0.751 &0.754\\
            \multicolumn{2}{l|}{} &$\pm$0.001 &$\pm$0.004 &$\pm$0.004 &$\pm$0.021 &$\pm$0.004 &$\pm$0.002 &$\pm$0.177 &$\pm$0.001 &$\pm$0.002 &$\pm$0.002\\
            \hline
             \multicolumn{2}{l|}{\multirow{2}{*}{Gradformer \cite{pei_Gradformer-Framework-MultiAspect_2024}}} &0.079 &0.646 &0.598 &0.334 &0.614 &0.732 &0.318 &0.769 &0.767 &0.756\\
            \multicolumn{2}{l|}{} &$\pm$0.001 &$\pm$0.004 &$\pm$0.006 &$\pm$0.013 &$\pm$0.006 &$\pm$0.005 &$\pm$0.139 &$\pm$0.006 &$\pm$0.004 &$\pm$0.003\\
            \hline
            \multicolumn{2}{l|}{\multirow{2}{*}{HIA (Ours)}} &\textbf{0.076} &\textbf{0.657} &\textbf{0.613} &\textbf{0.436} &\textbf{0.628} &\textbf{0.743} &\textbf{0.354} &\textbf{0.778} &\textbf{0.784} &0.764\\
            \multicolumn{2}{l|}{} &$\pm$\textbf{0.001} &$\pm$\textbf{0.004} &$\pm$\textbf{0.003} &$\pm$\textbf{0.043} &$\pm$\textbf{0.005} &$\pm$\textbf{0.002} &$\pm$\textbf{0.131} &$\pm$\textbf{0.006} &$\pm$\textbf{0.003} &$\pm$0.002\\
            \hline
        \end{tabular}
        \caption{The results of HIA and compared baselines on various pronunciation assessment tasks with average MSE (phoneme level) and PCC (phoneme, word, and utterance level) scores and standard deviations of five different runs. \label{table1}}
\end{table*}

\subsubsection{Word-level Modeling}
There is a high correlation between phoneme level and word level, so we leverage phoneme-level scores to calculate word-level scores. Specifically, we first sum the output $X$ of the Transformer encoder, the phoneme-level scoring results $S^{phn}$, and the word-level attention head ${H^{word}}$ as word-level inputs:
\begin{equation}
X^{word} = X + S^{phn} + H^{word}.
\end{equation}

There are many different aspects of word-level granularity, and scores of multiple aspects are related to each other and affect each other. In Word-level Modeling, we use the aspect attention mechanism \cite{do_Hierarchical-Pronunciation-Assessment_2023,ridley_Automated-Crossprompt-Scoring_2021} to capture the correlations between different aspects of the same granularity as well as the difference between the different aspects of scoring:
\begin{equation}
S^{word} = AspectAttn(X^{word}).
\end{equation}



Similar to the phoneme level, we add convolutional layer and regression heads to output the final accuracy, stress, and total score.

\subsubsection{Utterance-level Modeling}
The Transformer decoder is only used at the utterance level because the utterance level involves complex contextual information, and the decoder can capture long-range dependencies and global features \cite{pei_Gradformer-Framework-MultiAspect_2024}. Therefore, we use the decoupling method to model the utterance-level scoring task. First, we initialize a set of learnable vectors as queries, $Q^{utt}=\{{q_{k}^{utt}}\}_{k=1}^N$, $N$ is the number of utterance-level aspects. Then, the word-level scoring results $S^{word}$, the output $X$ of the Transformer encoder, and the utterance-level attention head $H^{utt}$ in the interactive attention module are summed up as the key and value into the Transformer decoder. The formula is defined as follows:
\begin{equation}
X^{utt} = X + S^{word} + H^{utt},
\end{equation}
\begin{equation}
S^{utt} = TransDecoder(Q^{utt}, X^{utt}).
\end{equation}

Finally, the output of Transformer decoder is first processed by convolutional layer and regression heads are added to predict the final utterance-level scores.

\subsection{Loss Function}
In this work, we use mean squared error (MSE) loss as loss function, which is widely used for pronunciation assessment. The formula is as follows:
\begin{equation}
\label{mse}
L_{MSE} = \frac{1}{N} \sum_{i=1}^{N} ({s_{i}} - {y_i})^{2},
\end{equation}
where ${N}$ is the number of samples,  ${s_{i}}$ is the ${i}$-th prediction score of the model, and  ${y_i}$ is the ${i}$-th ground truth.

As the reason of multi-aspect and multi-granularity pronunciation assessment task, we consider the total loss is calculated as the sum of each granularity level loss, and the loss at each granularity level is an average sum of corresponding multiple aspects:
\begin{equation}
L_{total} = \sum_{i=1}^{M}\frac{1}{N}\sum_{j=1}^{N}L_{ij},
\end{equation}
here ${M}$ and ${N}$ refer to the total number of granularity levels and corresponding aspect levels, respectively.

\begin{table*}[ht]
    \renewcommand{\arraystretch}{1.1} 
    \centering
        \begin{tabular}{c c|c c|c c c|c c c c c}
            \hline
            \multicolumn{2}{c|}{\multirow{2}{*}{\textbf{Model}}} &\multicolumn{2}{c|}{\textbf{Phoneme score}} &\multicolumn{3}{c|}{\textbf{Word score (PCC)}} &\multicolumn{5}{c}{\textbf{Utterance score (PCC)}} \\
            \cline{3-12}
            \multicolumn{2}{c|}{} &MSE$\downarrow$ &PCC$\uparrow$ &Acc$\uparrow$ &Stress$\uparrow$ &Total$\uparrow$ &Acc$\uparrow$ &Comp$\uparrow$ &Fluency$\uparrow$ &Prosodic$\uparrow$ &Total$\uparrow$ \\
            \hline
            \multicolumn{2}{l|}{\multirow{2}{*}{\makecell[c]{w/o P w/o W \\w/o U}}} &0.085 &0.626 &0.551 &0.335 &0.605 &0.717 &0.272 &0.751 &0.754 &0.748\\
            \multicolumn{2}{l|}{} &$\pm$0.000 &$\pm$0.006 &$\pm$0.004 &$\pm$0.020 &$\pm$0.006 &$\pm$0.003 &$\pm$0.159 &$\pm$0.003 &$\pm$0.003 &$\pm$0.004\\
            \hline
            \multicolumn{2}{l|}{\multirow{2}{*}{\makecell[c]{w/o P \textbf{w/} W \\\textbf{w/} U}}} &0.083 &0.621 &0.606 &0.429 &0.617 &0.737 &0.344 &0.765 &0.765 &0.758\\
            \multicolumn{2}{l|}{} &$\pm$0.001 &$\pm$0.005 &$\pm$0.006 &$\pm$0.024 &$\pm$0.007 &$\pm$0.005 &$\pm$0.118 &$\pm$0.004 &$\pm$0.005 &$\pm$0.003\\
            \hline
            \multicolumn{2}{l|}{\multirow{2}{*}{\makecell[c]{\textbf{w/} P w/o W \\\textbf{w/} U}}} &0.079 &\textbf{0.661} &0.569 &0.328 &0.604 &0.734 &0.322 &0.765 &0.771 &0.759\\
            \multicolumn{2}{l|}{} &$\pm$0.000 &$\pm$\textbf{0.005} &$\pm$0.004 &$\pm$0.023 &$\pm$0.006 &$\pm$0.002 &$\pm$0.105 &$\pm$0.005 &$\pm$0.004 &$\pm$0.004\\
            \hline
            \multicolumn{2}{l|}{\multirow{2}{*}{\makecell[c]{\textbf{w/} P \textbf{w/} W \\w/o U}}} &0.080 &0.653 &\textbf{0.615} &0.421 &0.621 &0.723 &0.302 &0.754 &0.753 &0.754\\
            \multicolumn{2}{l|}{} &$\pm$0.001 &$\pm$0.004 &$\pm$\textbf{0.003} &$\pm$0.011 &$\pm$0.006 &$\pm$0.001 &$\pm$0.074 &$\pm$0.003 &$\pm$0.003 &$\pm$0.002\\
            \hline
            \multicolumn{2}{l|}{\multirow{2}{*}{\makecell[c]{\textbf{w/} P \textbf{w/} W \\\textbf{w/} U \textbf{(HIA)}}}} &\textbf{0.076} &0.657 &0.613 &\textbf{0.436} &\textbf{0.628} &\textbf{0.743} &\textbf{0.354} &\textbf{0.778} &\textbf{0.784} &\textbf{0.764}\\
            \multicolumn{2}{l|}{} &$\pm$\textbf{0.001} &$\pm$0.004 &$\pm$0.006 &$\pm$\textbf{0.043} &$\pm$\textbf{0.007} &$\pm$\textbf{0.002} &$\pm$\textbf{0.151} &$\pm$\textbf{0.003} &$\pm$\textbf{0.004} &$\pm$\textbf{0.002}\\
            \hline
        \end{tabular}
        \caption{Ablation results on the effectiveness of Interactive Attention Module. P, W and U denote the interactive attention heads generated by the interaction attention module at the phoneme-, word-, and utterance-level granularity, respectively. \label{table2}}
\end{table*}

\section{Experiments}
\label{section4}
\subsection{Dataset}
Speechocean762 \cite{zhang_Speechocean762-OpenSource-NonNative_2021a}, currently the only open-source standard dataset designed specially for pronunciation assessment in read-aloud scenario, is used for our experiments. It consists of 5000 English sentences and the recorders are 250 non-native English speakers, half of whom are children. 

In addition, this dataset has a rich variety of data annotation types, independently annotated by five experts at the phoneme, word, and utterance levels. Specifically, for each utterance, it provides five utterance-level aspect scores: accuracy, fluency, completeness, prosody, and total score (ranging from 0-10). For each word, it provides three word-level aspect scores: accuracy, stress, and total score (ranging from 0-10). For each phoneme, it also provides an accuracy score(ranging from 0-2). In the experiments, the scores for word and utterance are uniformly rescaled to (0-2), making them on the same scale as the phoneme scores.

\subsection{Evaluation Metrics}
We use MSE to measure the difference between predicted scores and truth scores for phoneme level, the formula is shown in Eq. (\ref{mse}).

Pearson Correlation Coefficient (PCC) is also used as evaluation metric to measure the correlation of predicted values and labeled values of different aspects at each granularity level, it can be calculated as follows:
\begin{equation}
PCC({S}, {Y}) = \frac{\sum_{i=1}^{N}(s_{i}-\overline{s})(y_{i}-\overline{y})}{\sqrt{\sum_{i=1}^{N}(s_{i}-\overline{s})^2}\sqrt{{\sum_{i=1}^{N}}(y_{i}-\overline{y})^2}},
\end{equation}
where ${s_i}$, ${y_i}$ are the ${i}$-th predicted score given by our proposed model and corresponding true score given by the experts, respectively, $N$ is the total number of sentences.

\begin{table}[t]
    \renewcommand{\arraystretch}{1.1} 
    \centering
        \begin{tabular}{c c c c c c}
            \hline
            \multicolumn{2}{c}{\multirow{2}{*}{\textbf{Model}}} &\multirow{2}{*}{\textbf{Phoneme}} &\multirow{2}{*}{\textbf{Stress}} &\multirow{2}{*}{\textbf{Word}} &\multirow{2}{*}{\textbf{Utterance}} \\
            \multicolumn{2}{c}{} & & & & \\
            \hline
            \multicolumn{2}{l}{\multirow{2}{*}{HIA}} &\textbf{0.657} &\textbf{0.436} &\textbf{0.628} &\textbf{0.764} \\
            \multicolumn{2}{l}{} &\textbf{$\pm$0.004} &\textbf{$\pm$0.043} &\textbf{$\pm$0.007} &\textbf{$\pm$0.002} \\

            \multicolumn{2}{l}{\multirow{2}{*}{--Res}} &0.647 &0.382 &0.603 &0.748 \\
            \multicolumn{2}{l}{}  &$\pm$0.007 &$\pm$0.021 &$\pm$0.009 &$\pm$0.003 \\

            \multicolumn{2}{l}{\multirow{2}{*}{--Hi}} &0.645 &0.374 &0.593 &0.753 \\
            \multicolumn{2}{l}{} &$\pm$0.001 &$\pm$0.016 &$\pm$0.001 &$\pm$0.003 \\
            
            \hline
        \end{tabular}
        \caption{Ablation results on the effectiveness of Residual Hierarchical structure. Res denotes residual structure, Hi denotes hierarchical structure. Because of the space limitation, only the PCC of phoneme accuracy, word stress, word total and utterance total scores are reported.\label{table3}}
\end{table}

\subsection{Experimental Setup}
\subsubsection{Training Configuration} For the model training phase, we use Adam optimizer to train the HIA and initialize the learning rate to 1e-3, the learning rate is halved every 5 epochs after the 20th epoch. The maximum number of epoch is set to 100 and we save the model with the minimum phoneme-level MSE loss as the optimal model. For all experiments, we perform five times with different random seeds for all models, whose mean and standard deviation are reported.

\subsubsection{Model Configuration} In HIA, the layers of Transformer encoder and decoder are set to 3 and their embedding dimensions are 48. For dimension matching, the embedding dimension of interactive attention module query is also set to 48. Due to the dataset and feature dimension are not large enough, we set the number of heads for self-attention and cross-attention in interactive attention module and Transformer to 1. The dropout ratio is set to 0.1 to suppress overfitting. In addition, the kernel size of convolutional layers is set to 5 for each granularity and stride is set to 1.

\section{Results and Discussions}
\label{section5}
\subsection{Main Results}
In this section, we compare our proposed HIA with traditional single-granularity scoring models and state-of-the-art multi-aspect multi-granularity scoring baseline models, all baseline results are quoted from their original papers and summarized in Table \ref{table1}. According to the results, we have the following observations:
\begin{itemize}
    \item Our model outperforms the evaluation results of human experts in all but the utterance-level completeness metric. This gap is mainly attributed to the distributional bias in the dataset, in which 4975 out of 5000 sentences in the dataset have completeness scores of 10.
    \item Compared with the single-granularity scoring methods, HIA demonstrates significant performance advantages in all metrics except the total score at utterance level. This suggests that the multi-aspect multi-granularity scoring approach can better utilize the different inter-granularity correlations and dependencies in audio data.
    \item Compared with multi-aspect multi-granularity scoring baseline models, our model consistently achieved the state-of-the-art results. Its highest PCC scores highlights the ability of HIA to handle complex pronunciation features, and demonstrates our proposed model is capable of processing and evaluating articulatory features of different granularities more effectively.
\end{itemize}



\subsection{Ablation Studies}
In order to delve deeper into the key factors that enhance the effectiveness of HIA, we conduct ablation experiments to study the effects of the interactive attention module, the residual hierarchical structure, the number of convolutional layers and the model Configuration on model performance.

\begin{table}[t]
    \renewcommand{\arraystretch}{1.1} 
    \centering
        \begin{tabular}{c c c c c c}
            \hline
            \multicolumn{2}{c }{\multirow{2}{*}{\textbf{Layer}}} &\multirow{2}{*}{\textbf{Phoneme}} &\multirow{2}{*}{\textbf{Stress}} &\multirow{2}{*}{\textbf{Word}} &\multirow{2}{*}{\textbf{Utterance}} \\
            \multicolumn{2}{c}{} & & & & \\
            \hline
            \multicolumn{2}{l}{\multirow{2}{*}{0 layer}} &0.638 &0.415 &0.601 &0.754 \\
            \multicolumn{2}{l}{} &$\pm$0.007 &$\pm$0.011 &$\pm$0.012 &$\pm$0.003 \\
            \hline
            \multicolumn{2}{l}{\multirow{2}{*}{1 layer*}} &\textbf{0.657} &\textbf{0.436} &\textbf{0.628} &\textbf{0.764} \\
            \multicolumn{2}{l}{} &\textbf{$\pm$0.004} &\textbf{$\pm$0.043} &\textbf{$\pm$0.007} &\textbf{$\pm$0.002} \\
            \hline
            \multicolumn{2}{l}{\multirow{2}{*}{2 layers}} &0.646 &0.427 &0.618 &0.759 \\
            \multicolumn{2}{l}{} &$\pm$0.002 &$\pm$0.023 &$\pm$0.004 &$\pm$0.004 \\
            \hline
           \multicolumn{2}{l}{\multirow{2}{*}{3 layers}} &0.645 &0.421 &0.617 &0.755 \\
            \multicolumn{2}{l}{} &$\pm$0.008 &$\pm$0.007 &$\pm$0.013 &$\pm$0.005 \\
            \hline
        \end{tabular}
        \caption{Ablation results on the effectiveness of the number of convolutional layers. * denotes the setting used in HIA model.\label{table4}}
\end{table}

\begin{table}[t]
    \renewcommand{\arraystretch}{1.1} 
    \centering
        \begin{tabular}{c c c c c c}
            \hline
            \multicolumn{2}{c}{\multirow{2}{*}{\textbf{Setting}}} &\multirow{2}{*}{\textbf{Phoneme}} &\multirow{2}{*}{\textbf{Stress}} &\multirow{2}{*}{\textbf{Word}} &\multirow{2}{*}{\textbf{Utterance}} \\
            \multicolumn{2}{c}{} & & & & \\
            
            \hline
            \multicolumn{6}{l}{\textit{Embedding Size}} \\
            
            \hline
            \multicolumn{2}{c}{24} &0.649 &0.420 &0.613 &0.752 \\

            \multicolumn{2}{c}{48*} &\textbf{0.657} &\textbf{0.436} &\textbf{0.628} &\textbf{0.764} \\

            \multicolumn{2}{c}{96} &0.654 &0.432 &0.611 &0.762 \\
            
            \hline
            \multicolumn{6}{l}{\textit{Number of Heads}} \\
            
            \hline
            \multicolumn{2}{c}{1*} &\textbf{0.657} &\textbf{0.436} &\textbf{0.628} &\textbf{0.764} \\
            
            \multicolumn{2}{c}{2} &0.652 &0.431 &0.618 &0.759 \\
            
            \multicolumn{2}{c}{4} &0.648 &0.433 &0.623 &0.751 \\
            
            
            \hline
        \end{tabular}
        \caption{Ablation results on different model configuration. * denotes the setting used in HIA model.\label{table5}}
\end{table}


\subsubsection{Interactive Attention Module Ablation} To validate the effectiveness of the interactive attention module, we conduct ablation study on interactive attention heads at each granularity level, and the results are shown in Table \ref{table2}. The first row represents the ablation of all granularity interactive attention heads, with only the residual hierarchical structure used to score each granularity.

It can be seen that using the corresponding interactive attention heads at each granularity level (rows 2 to 4) improves the performance of metrics at each granularity level, demonstrating the interactive attention module benefits each granularity level. In particular, using word-level attention heads significantly improves performance on word stress, validating the correctness of using the interactive attention module for bidirectional interaction modeling. Using interactive attention heads at all granularity levels (row 5) achieves the best performance, further confirming that the interactive attention module can effectively capture the interdependencies between different granularity levels.

\subsubsection{Residual Hierarchical Structure Ablation} As shown in Table \ref{table3}, after ablating the residual connection, all the metrics decline to some extent, especially for the word stress scores, validating the residual structure affords overall performance improvements. The hierarchical structure is embodied in score passes between adjacent granularities, so we ablate the hierarchical structure by removing score passes. It can be seen that the removal of the hierarchical structure also resulted in decreases on all metrics, which further confirms the importance of the hierarchical structure in improving model performance.



\subsubsection{Convolutional Layer Ablation:} To address the neglect of local context clues that may result from the feature extraction process, we introduce convolutional layer to enhance the model's ability to capture local features. Table \ref{table4} shows that compared with not using convolutional layers (0 layer), HIA achieves performance improvements on all pronunciation assessment metrics with the introduction of convolutional layers. It begins to decline with 2 and 3 convolutional layers, because the the dataset is not large enough and the increased parameters are difficult to optimize.

\subsubsection{Model Size Ablation:} To investigate the impact of model capacity on performance and assess the scalability of HIA, we conduct ablation studies on two configuration parameters: embedding size and number of attention heads. As shown in Table \ref{table5}, increasing the embedding size from 24 to 48 leads to consistent improvements across all metrics. However, further increasing the embedding size results in minor performance drops.


Similarly, we can observe slightly lower results with multiple heads, which we attribute to the limited data size and increased model complexity making optimization more difficult. These findings indicate that the selected configuration strikes a balance between expressiveness and efficiency. 

\subsection{Data Correlation Analysis}
To validate the high correlation between phoneme-, word- and utterance-level scores, we calculate the correlation between each pair of aspects and visualize it.

\begin{figure}[t]
\centering
\includegraphics[width=3.1 in]{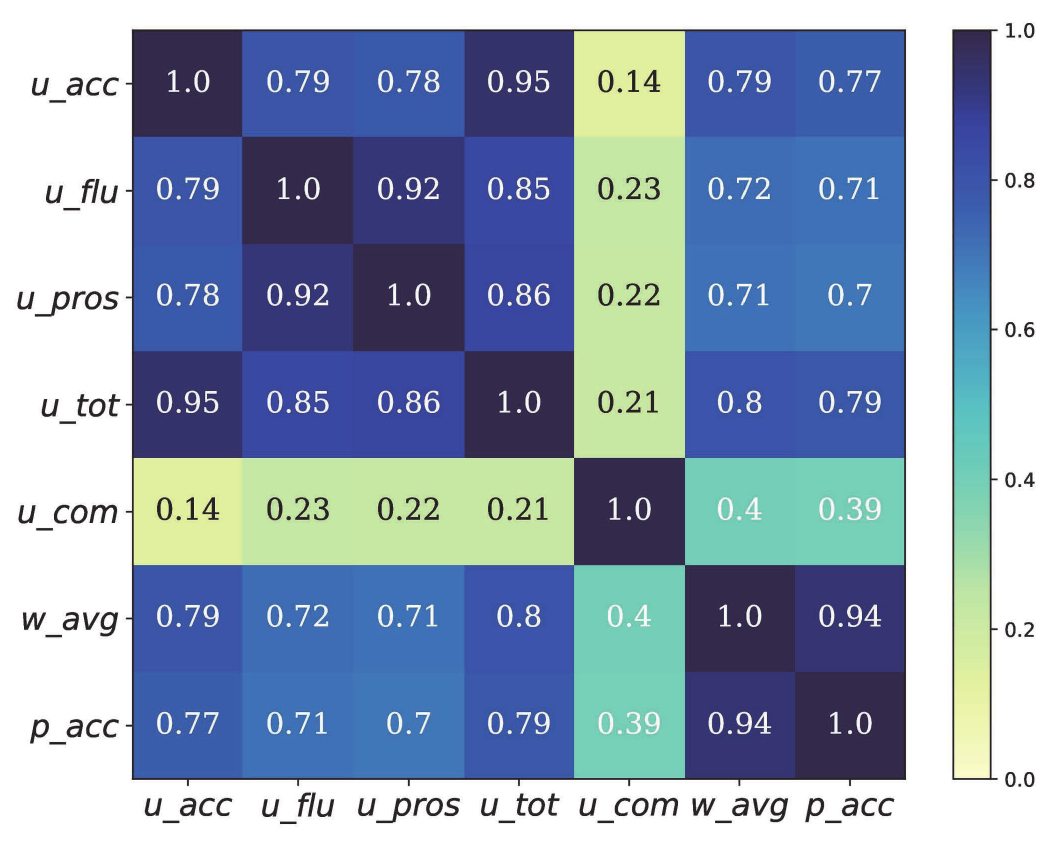}
\caption{Correlation matrix of different metrics at three granularities. Thereinto, p\_acc stands for phoneme-level accuracy; w\_avg stands the mean value for word-level accuracy, total and stress; u\_com, u\_acc, u\_flu, u\_pros and u\_tot stand for utterance-level completeness, accuracy, fluency, prosodic and total score, respectively.}
\label{fig_4}
\end{figure}

As shown in Fig. \ref{fig_4}, the relatively high correlations among phoneme accuracy, word mean scores, and utterance-level scores suggests that these scores are interdependent. This observation supports the use of bidirectional interaction mechanisms and hierarchical structures for modeling linguistic structures and accomplishing multi-granularity scoring tasks.

\section{Conclusion}
\label{section6}
In this paper, we propose a novel multi-aspect multi-granular pronunciation assessment model named HIA. To achieve bidirectional interaction between different granularity levels, we design a novel Interactive Attention Module that generates interactive attention heads corresponding to each granularity, significantly improves the model performance, particularly on word stress. In addition, we propose a residual hierarchical structure through residual connections to mitigate feature forgetting, further improving the model performance. Experimental results on speechocean762 dataset show that our proposed model achieves the state-of-the-art of the multi-aspect multi-granularity pronunciation assessment on all granularities and aspects metrics.

\appendix
\bigskip

\section{Acknowledgments}
\label{section7}
This work was supported in part by the National Natural Science Foundation of China under Grant  62172256, 62202272, 62202278, in part by Natural Science Foundation of Shandong Province under Grant ZR2024LZH002 and Taishan Scholar Project of Shandong Province under Grant tstp20250704.

\bibliography{ref}

\end{document}